\documentclass[a4paper,twoside]{article}

\usepackage[]{natbib}

\usepackage{epsfig}
\usepackage{subcaption}
\usepackage{calc}
\usepackage{amssymb}
\usepackage{amstext}
\usepackage{amsmath}
\usepackage{amsthm}
\usepackage{multicol}
\usepackage{pslatex}
\usepackage{apalike}
\usepackage{algorithm2e}
\usepackage[bottom]{footmisc}

\usepackage{booktabs}
\usepackage{siunitx}
\usepackage[capitalize]{cleveref}
\usepackage{url}

\usepackage{SCITEPRESS}     %

\begin{document}

\title{UCorr: Wire Detection and Depth Estimation for Autonomous Drones}

\author{\authorname{Benedikt Kolbeinsson and Krystian Mikolajczyk}
\affiliation{Imperial College London}
\email{bk915@imperial.ac.uk, k.mikolajczyk@imperial.ac.uk}
}


\abstract{
In the realm of fully autonomous drones, the accurate detection of obstacles is paramount to ensure safe navigation and prevent collisions. Among these challenges, the detection of wires stands out due to their slender profile, which poses a unique and intricate problem.
To address this issue, we present an innovative solution in the form of a monocular end-to-end model for wire segmentation and depth estimation. Our approach leverages a temporal correlation layer trained on synthetic data, providing the model with the ability to effectively tackle the complex joint task of wire detection and depth estimation.
We demonstrate the superiority of our proposed method over existing competitive approaches in the joint task of wire detection and depth estimation. Our results underscore the potential of our model to enhance the safety and precision of autonomous drones, shedding light on its promising applications in real-world scenarios.}

\onecolumn \maketitle \normalsize \setcounter{footnote}{0} \vfill

\section{\uppercase{INTRODUCTION}}
\label{sec:introduction}

In the era of autonomous systems and unmanned aerial vehicles (UAVs), the ability to navigate through complex environments with precision and safety is of paramount importance. One critical aspect of this challenge is the accurate detection of obstacles, a task that holds the key to preventing collisions and ensuring successful mission execution. Among these potential obstacles, wires, with their slim and inconspicuous profiles, represent a particularly formidable challenge. In this research, we delve into the intricate world of wire detection and depth estimation, unveiling a novel and effective approach that holds promise for enhancing the capabilities of autonomous drones in real-world scenarios.

 UAVs have evolved to encompass a wide spectrum of capabilities, from those under remote human control to fully autonomous systems. UAVs have found extensive applications, including forestry research \citep{tang_drone_2015}, autonomous inspections of electrical distribution networks \citep{nguyen_automatic_2018} and package delivery \citep{benarbia2021literature}. Notably, the utilization of UAVs in disaster response operations has garnered significant attention due to their potential critical roles \citep{erdelj_help_2017,adams2011survey,estrada2019uses,pi2020convolutional,daud2022applications,qu2023environmentally}. These versatile aerial platforms offer promising solutions for various real-world challenges, setting the stage for innovations that can enhance safety, efficiency, and effectiveness in diverse domains.
 
 In the pursuit of safe and collision-free flight, UAVs have traditionally relied on obstacle detection systems, often employing proximity sensors based on ultrasound or computer vision. However, these existing systems face a notable limitation: their inability to consistently and reliably detect thin obstacles, such as power lines, telephone wires, and structural cables.

The weight of a UAV is a critical factor that directly influences its efficiency and maneuverability. Integrating additional sensors, such as LiDAR, for improved object detection can yield significant benefits but often comes at the cost of undesirable trade-offs. These compromises include adverse effects on flight characteristics and heightened expenditure. In this context, it is worth noting that nearly all UAVs are equipped with cameras for a multitude of purposes. Leveraging these onboard cameras to detect wires and obstacles not only eliminates the need for additional weight but also circumvents the burden of extra hardware costs.

Detecting wires in images presents a formidable challenge stemming from multiple factors. Wires possess inherent thinness, often manifesting as single-pixel or sub-pixel entities. Their subtle presence can seamlessly blend into complex and cluttered backgrounds, rendering them elusive even to human observers. With limited distinctive features, wires at the pixel level can bear a striking resemblance to other commonplace structures. However, merely identifying wires within an image is insufficient. Crucially, gauging the distance to these obstacles is paramount, as closer objects pose a heightened risk compared to distant ones. Moreover, to enable intelligent navigation through its environment, a drone must establish a comprehensive understanding of its surroundings

To tackle these challenges, we introduce UCorr, a monocular wire segmentation and depth estimation model. Utilizing a temporal correlation layer within an encoder-decoder architecture, as illustrated in \Cref{fig:ucorr_schematic}, our approach surpasses the performance of existing methods in the domain of wire detection and depth estimation.

In summary, our contributions are as follows:
\begin{itemize}
    \item We present UCorr, an innovative model tailored for monocular wire segmentation and depth estimation.
    \item We demonstrate that UCorr outperforms current methods, showcasing its potential to advance wire detection and depth estimation in autonomous systems.
    \item We introduce a novel wire depth evaluation metric designed to accurately assess wire depth estimation. This metric accounts for the unique characteristics of wires, providing a tailored and comprehensive evaluation.
\end{itemize}

\section{RELATED WORK}
In this section, we provide an overview of related work on wire detection and depth estimation, discussing them separately due to limited research on their joint task.

\subsection{Wire Detection}
Academic research has predominantly concentrated on wire detection in images, with comparatively less emphasis placed on addressing the challenge of accurately determining the distance to the wires.

\paragraph{Traditional computer vision techniques.}

Early work \citep{kasturi2002wire}, proposed using the Steger algorithm \citep{steger1998unbiased} to detect edges on real images with synthetic wires, followed by a thresholded Hough transform \citep{duda_use_1972}. 
This quickly became the standard approach for wire detection and following work used variations of these three stages: (1) An edge detector, (2) the Hough transform and finally (3) a filter.

For example, \cite{li_knowledge-based_2008} first use a Pulse-Coupled Neural Network (PCNN) to filter the background of the images before using the Hough transform to detect straight lines. Then, using k-means clustering, power lines are detected and other line-like objects discarded. 
Similarly, \cite{sanders-reed_passive_2009} begin by removing large clutter using a Ring Median Filter and a SUSAN filter. To find wire like segments they use a gradient phase operator and vector path integration. Then merge small line segments together using morphological filters. Lastly, temporal information is used to remove non persistent line segments to reduce the false alarm rate.
\cite{zhang_high_2012} start by using a gradient filter followed by the Hough transform to find line segments. Then k-means is used to select power lines and to discard other line-like objects. Lastly, using temporal information, the power lines are tracked using a Kalman filter.
\cite{candamo_detection_2009} combine temporal information to estimate pixel motion and a Canny edge detector \citep{canny1986computational} to form a feature map. This is followed by a windowed Hough transformation. The motion model is used to predict the next location of detected lines. 
\cite{song_power_2014} create an edge map using a matched filter and the first-order derivative of Gaussian.  Morphological filtering is used to detect line segments before a graph-cut model groups line segments into whole lines. A final morphological filter is applied again to remove false lines.
A slightly different approach was taken by \cite{zhou_fast_2017} where they developed two methods. The first one, for a monocular camera which requires an inertial measurement unit and a second one, a stereo camera solution. Both start with a DoG edge detector \citep{marr1980theory} to detect edge points before reconstructing them in 3D space using temporal information.

\paragraph{Deep learning techniques.}
More recently, deep learning techniques have become more popular.
\cite{lee_weakly_2017} propose a weakly supervised CNN where the training images only have class labels. Multiple feature maps are generated at different depths of the network and are scaled and merged together to produce a final mask.
\cite{madaan_wire_2017} propose multiple variations of dilated convolutional neural networks (DCNN) trained on both synthetic data and real data.
\cite{stambler_detection_2019} use a CNN for feature generation then using two separate CNN networks, one of which classifies whether a wire is located near an anchor point while the other produces a Hesse norm line from the anchor to the detected wire. A Kalman filter helps tack the wires between frames and the wire's relative location is calculated.
\cite{zhang_combined_2019} use an edge detector proposed by \cite{liu_richer_2017} which is a modified version of VGG16 \citep{simonyan_very_2014}. To remove the noise, only the longest edges with high confidence are kept.
\cite{nguyen_ls-net:_2020} use a CNN based on VGG16 to generate feature maps on different grids on the image. A classifier determines whether a wire appears on a grid and then a separate regressor network outputs the location of the longest line segment in each grid.

More recently, \cite{chiu2023automatic} propose a two-stage wire segmentation model where a coarse module focuses on capturing global contextual information and identifying regions potentially containing wires. Then a local module analyzes local wire-containing patches.

In contrast to these approaches, our model tackles wire detection and depth estimation simultaneously, trained end-to-end, and augmented with a correlation layer.

\subsection{Depth Estimation} \label{section:related-work_depth_estimation}
Monocular depth estimation is an ongoing research problem that has seen continuous improvements over time. The problem of monocular depth estimation is comprised of a single image of a scene with the goal of producing the depth values for that scene. Here we present a brief overview over methods developed specifically for depth estimation.

Convolutional neural networks are a common method used to tackle this problem. For example, \cite{eigen2014depth} propose a two stack approach. First a global network predicts a coarse depth map. The second local fine-scale network is applied to both the original image as well as the output of the global network, to produce the final output. Later \cite{eigen2015predicting} improve the network by making it deeper and add a third stage for higher resolution. Whereas, \cite{laina2016deeper} propose a fully convolutional network in an encoder-decoder setup. The network uses ResNet-50 \citep{he2016deep} as its backbone followed by unpooling and convolutional layers. They found that the reverse Huber loss, termed berHu \citep{owen2007robust}, a mix between the standard \(L_1\) and \(L_2\) losses, performed well.

\cite{godard2017unsupervised} present an unsupervised approach which uses image reconstruction loss as the training signal. In other words, instead of using the ground truth depth, which is difficult to acquire, they use a pair of binocular cameras (two cameras side-by-side) and learn to generate each image given its pair. Importantly, they compute both the left-to-right and right-to-left disparities using only the left input image. This allows them to achieve better depth prediction as both predictions should be consistent.
Following this, \cite{monodepth2} introduce an updated approach, Monodepth2, which uses a standard UNet \citep{ronneberger2015u} for depth prediction and a simple encoder to estimate the pose between images. By ignoring occluded pixels and pixels which violate motion assumption, they achieve greater performance.

More recently, methods \citep{bhat2021adabins,li2022binsformer,bhat2023zoedepth} utilize a binning technique where the model estimates continuous depth by combining predicted probability distributions and discrete bins through a linear process.

\begin{figure*}[t]
    \centering
    \includegraphics[width=\linewidth]{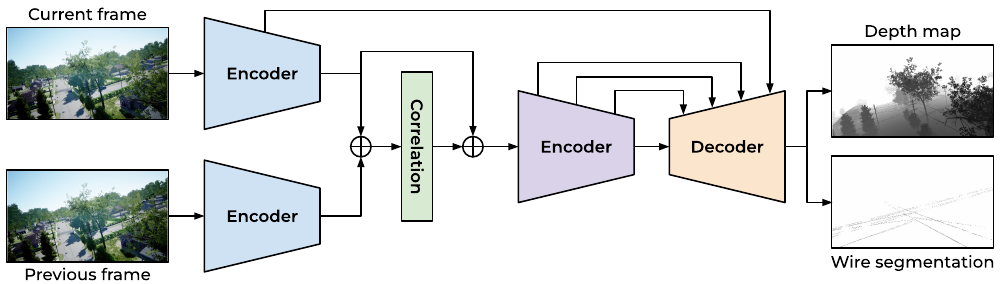}
    \caption{Schematic of UCorr. Two sequential frames are used as input (both RGB). The leftmost encoders share weights. The output consist of a binary wire semantic segmentation map for the target image along with a full depth map. See \Cref{sec:method} for more details.}
    \label{fig:ucorr_schematic}
\end{figure*}

\section{METHOD} \label{sec:method}
In this section, we present our method for wire detection and depth estimation. To begin, we will discuss the motivation and underlying principles that guided the development of our approach.

\subsection{Motivation}

From a visual perspective, wires have few unique visual features. The most obvious of which is their shape and color. One important feature common to most wires is their uniform construction. Note that significant differences still exist between wires, but individual wires will have consistent features, such as their width. However, an individual wire can be exposed to multiple different environmental factors across a scene resulting in perceived global differences. 

To exploit some of this inherent local consistency, and occasional global consistency, we propose a self correlation layer. A correlation layer, much like a convolutional layer, applies a kernel to an input image. Unlike a convolutional layer, the kernel in the correlation layer does not include any learned weights but instead consists of data. This data can be a second image and thus the output represents the correlation between the input image and the second image. The output tensor consists of the correlation between each pixel or patch from both images. 
Given two patches of size \(K = 2k + 1\), centered at \(\textbf{p}_1\) and \(\textbf{p}_2\), a single comparison between them can be defined as:

\begin{equation}
    c(\textbf{p}_1,\textbf{p}_2) = \sum_{\textbf{o} \in [-k,k] \times [-k,k]}\langle\,\textbf{f}_1(\textbf{p}_1 + \textbf{o}),\textbf{f}_2(\textbf{p}_2 + \textbf{o})\rangle
    \label{equation:correlation}
\end{equation}
where \(\textbf{f}_1\) and \(\textbf{f}_2\) and are multi-channel feature maps.
Our implementation is inspired on FlowNet \citep{dosovitskiy2015flownet} which also introduces a maximum displacement parameter. This prevents calculating the correlation of pixels in completely different parts of the images. Instead, our proposed method only calculates the local correlation around each patch. 

This correlation layer does not only benefit the wire segmentation capabilities of the model but also helps with depth estimation. Monocular depth estimation is inherently difficult due to the lack of three dimensional perspective. However, as a drone flies, multiple frames from the drone's camera can be recorded, which in turn provide rich temporal information. We hypothesize that the correlation layer helps extract this information as it allows the model to match objects between frames. Thus the flow of the scene can be understood as objects closer will have a larger displacement between frames compared to objects further away.

To fully utilize the correlation layer we propose UCorr, an end-to-end wire segmentation and depth estimation model. UCorr is the result of strategically adding a temporal correlation layer to the UNet \citep{ronneberger2015u} architecture. UNet, first developed for biomedical image segmentation, has now has become the \textit{de facto} baseline for all segmentation tasks.

\subsection{UCorr Network Architecture}

UCorr comprises two independent input paths, illustrated in \Cref{fig:ucorr_schematic}. The first path handles the current image frame from the drone's RGB camera, while the second path processes the previous image frame. These initial encoders have common architecture and weights. The encoder pair use a set of convolutional layers and max pooling to compose an encoding of the input frames. This design enables the correlation layer to correlated between learned features from each image frame rather than raw pixel values. The remainder of the network features a convolutional auto-encoder with skip connections to alleviate information bottlenecks, similar to UNet. Importantly, there is a skip-connection from the first encoder (the one with the current frame) to the decoder.

In addition, we explore variations of this architecture in our ablation studies in \Cref{subsec:ablation}.

\subsection{Loss Function}

We propose to use a loss function that incorporates both a wire segmentation component and a depth estimation component. The total loss function is defined as: 
\begin{equation}
    \mathcal{L}_{total} = \mathcal{L}_{wire} + \mathcal{L}_{depth}
\end{equation}
where \(\mathcal{L}_{wire}\) is the cross-entropy loss for pixel-wise binary classification of wires, thus:
\begin{equation}
    \mathcal{L}_{wire} = -w{(y\log(p) + (1 - y)\log(1 - p))}
\end{equation}
Where \(y\) is the binary label, \(p\) is the predicted label and \(w\) is an optional weight. Due to the class imbalance, the positive class is weighted 20 times the negative class in the loss. The depth loss has two components: 
\begin{equation}
    \mathcal{L}_{depth} = \mathcal{L}_{MAE} + \lambda\mathcal{L}_{MS-SSIM}
\end{equation}
Where \(\mathcal{L}_{MAE}\) is the pixel-wise mean absolute error and \(\mathcal{L}_{MS-SSIM}\) is the multi-scale structural similarity \citep{wang2003multiscale}. \(\lambda\) is set to 0.8.

\begin{table*}[t] \centering
\caption{Wire segmentation on our simulated flights. Models shown are: Canny \citep{canny1986computational}, DCNN \citep{madaan_wire_2017}, U-Net \citep{ronneberger2015u} and UCorr (ours). Best results shown in bold. Due to the large class imbalance (very few pixels of wires), metrics such as AUC can be misleading. However, UCorr outperforms all the other methods in every metric.}
\begin{small}
\begin{tabular}{@{}lccccccc@{}}\toprule
Model & Depth & IoU \((\uparrow)\) & AUC \((\uparrow)\) & AP \((\uparrow)\) & Precision \((\uparrow)\) & Recall \((\uparrow)\) & F1 \((\uparrow)\)\\ \midrule
Canny & No & 0.011 & - & - & 0.012 & 0.220 & 0.022\\ 
DCNN & No & 0.030 & 0.866 & 0.077 & 0.058 & 0.217 & 0.083\\ 
UNet & Yes & 0.123 & 0.986 & 0.419 & 0.219 & 0.589 & 0.307\\ 
UCorr (ours) & Yes & \textbf{0.138} & \textbf{0.989} & \textbf{0.451} & \textbf{0.247} & \textbf{0.605} & \textbf{0.339}\\ 
\bottomrule
\end{tabular}
\end{small}
\label{tab:model_comparison_synthetic_wire}
\end{table*}

\begin{table*}[t] \centering
\caption{Depth estimation on our simulated flights. Best results shown in bold. The Absolute Relative Error for Wire Depth (Abs. Rel. WD) is an especially challenging metric, which UCorr performs relatively well in.}
\begin{small}
\begin{tabular}{@{}lcccc@{}}\toprule
Model & Segmentation & abs. rel. \((\downarrow)\) & MAE \((\downarrow)\) & abs. rel. WD \((\downarrow)\)\\ \midrule
UNet & Yes & 0.129 & \textbf{3.414} & 0.606\\ 
UCorr (ours) & Yes & \textbf{0.128} & 3.701 & \textbf{0.564}\\  
\bottomrule
\end{tabular}
\end{small}
\label{tab:model_comparison_synthetic_depth}
\end{table*}

\begin{figure*}[t!]
\begin{center}
\begin{tabular}{@{}c@{\hspace{1.5mm}}c@{\hspace{1.5mm}}c@{\hspace{1.5mm}}c@{\hspace{1.5mm}}c@{}}
\includegraphics[width=0.1906\linewidth]{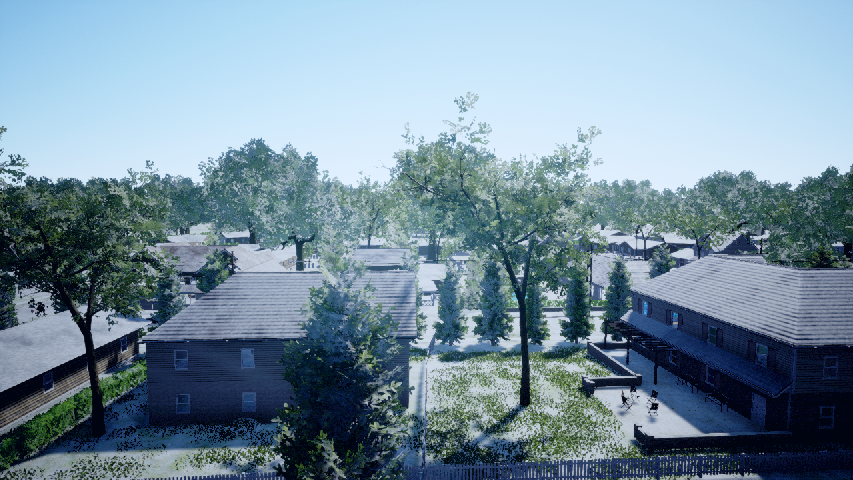}&
\includegraphics[width=0.1906\linewidth]{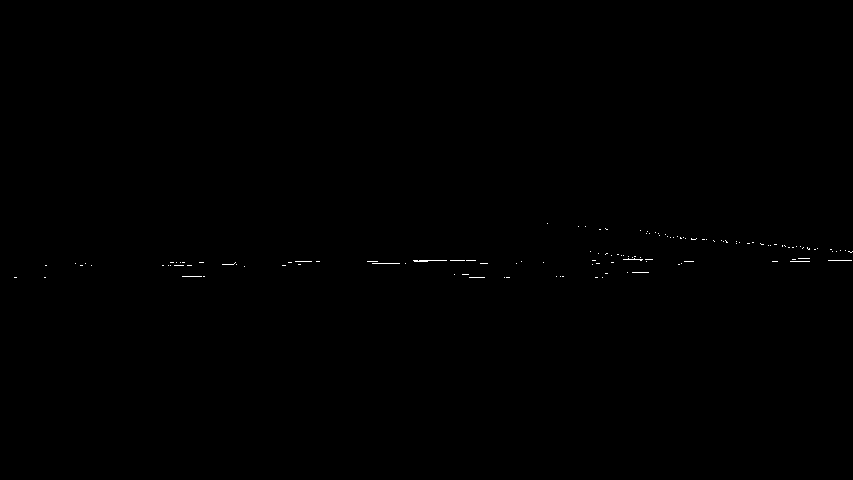}&
\includegraphics[width=0.1906\linewidth]{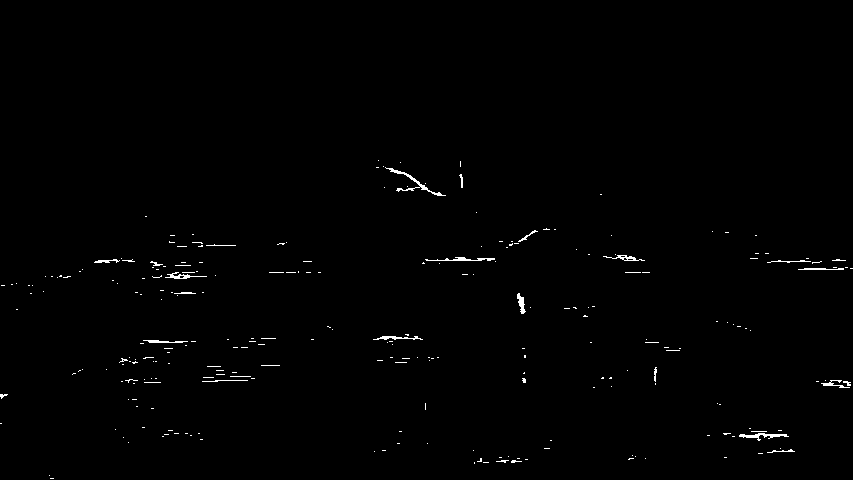}&
\includegraphics[width=0.1906\linewidth]{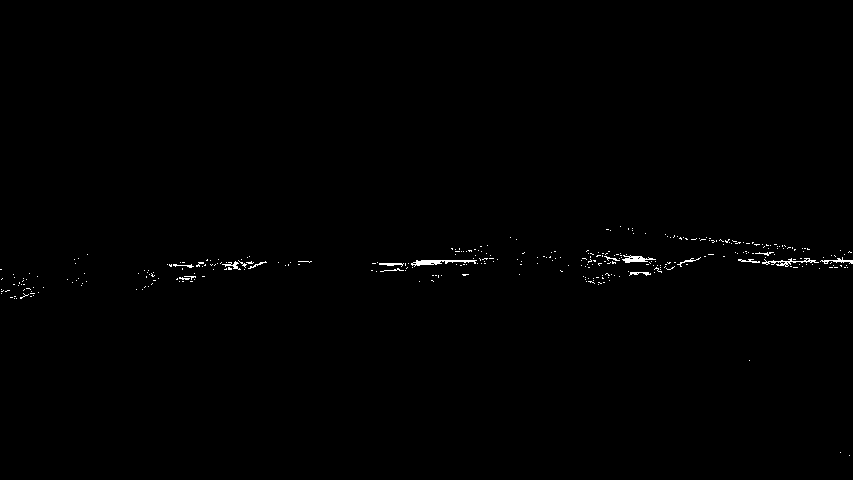}&
\includegraphics[width=0.1906\linewidth]{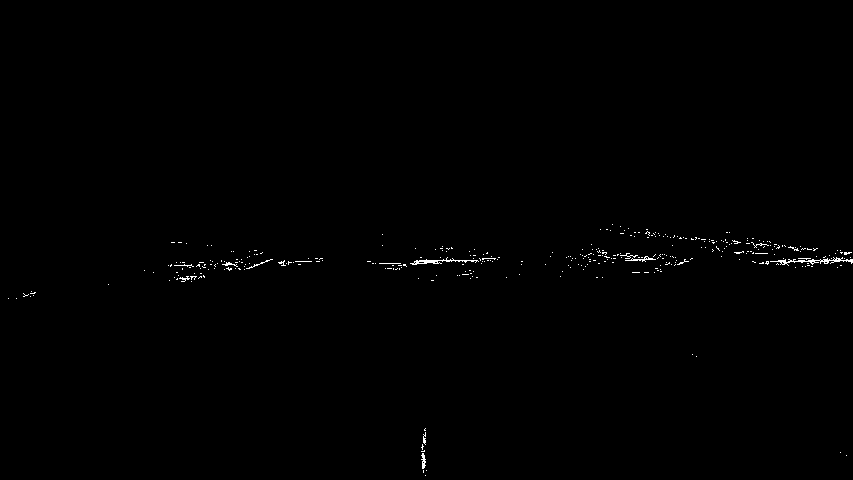}\\
\includegraphics[width=0.1906\linewidth]{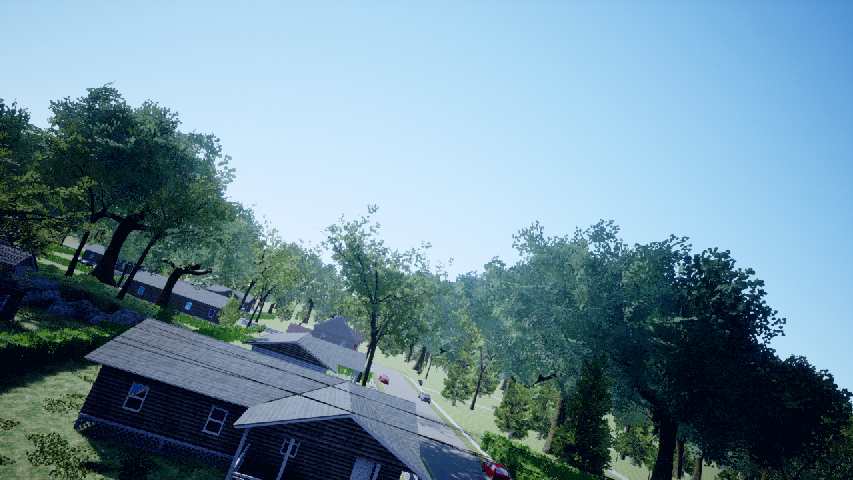}&
\includegraphics[width=0.1906\linewidth]{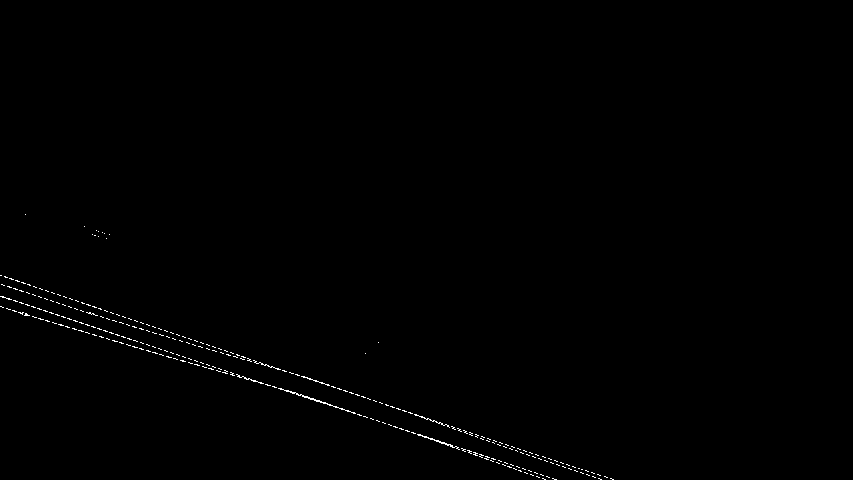}&
\includegraphics[width=0.1906\linewidth]{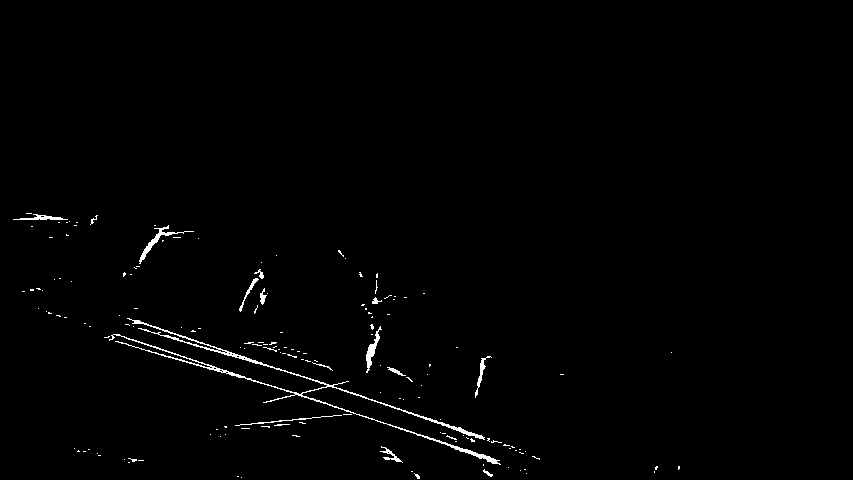}&
\includegraphics[width=0.1906\linewidth]{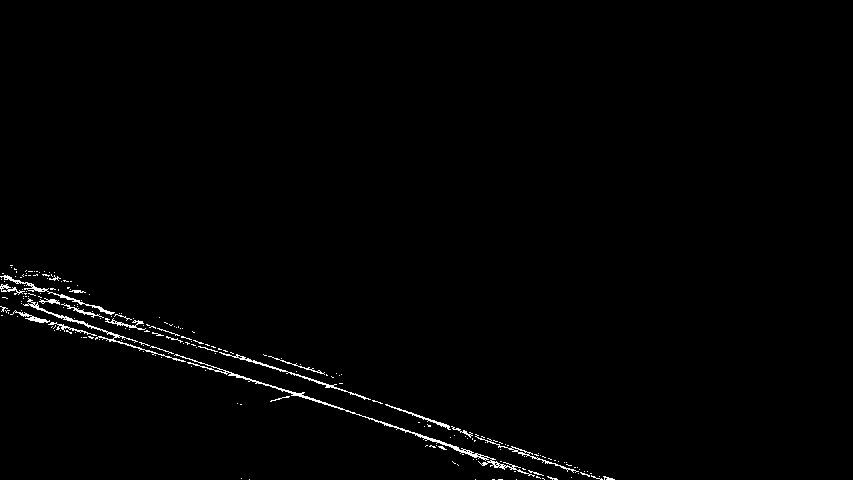}&
\includegraphics[width=0.1906\linewidth]{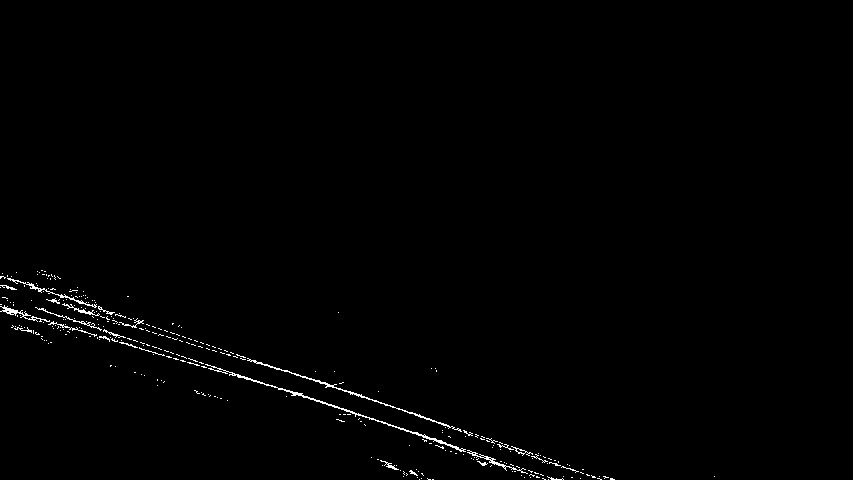}\\
\includegraphics[width=0.1906\linewidth]{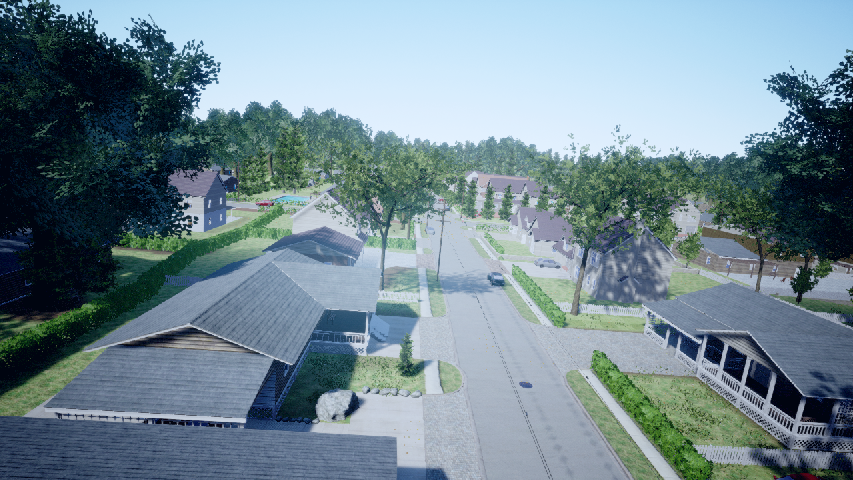}&
\includegraphics[width=0.1906\linewidth]{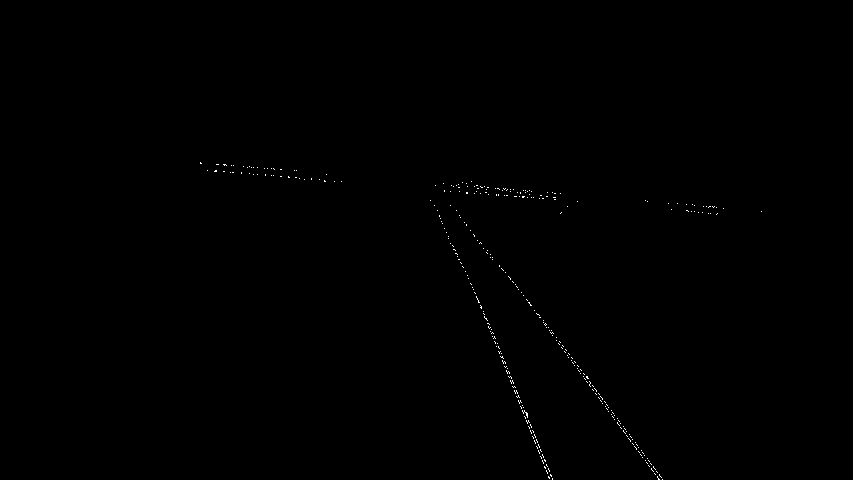}&
\includegraphics[width=0.1906\linewidth]{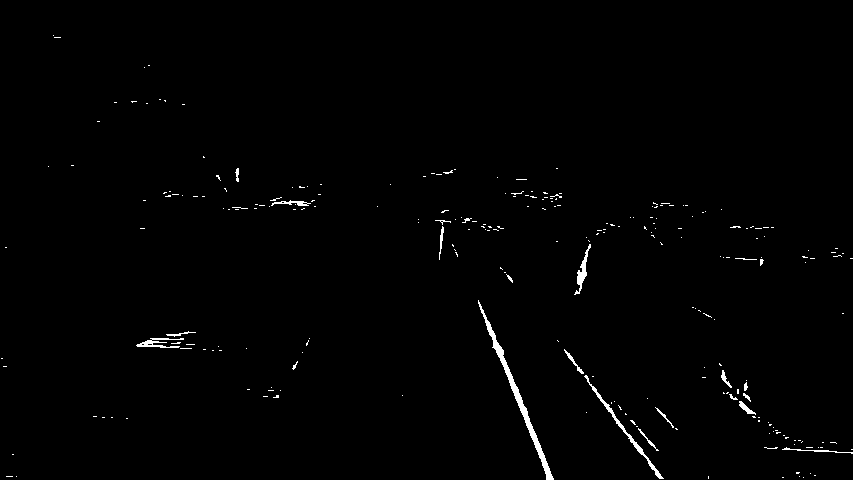}&
\includegraphics[width=0.1906\linewidth]{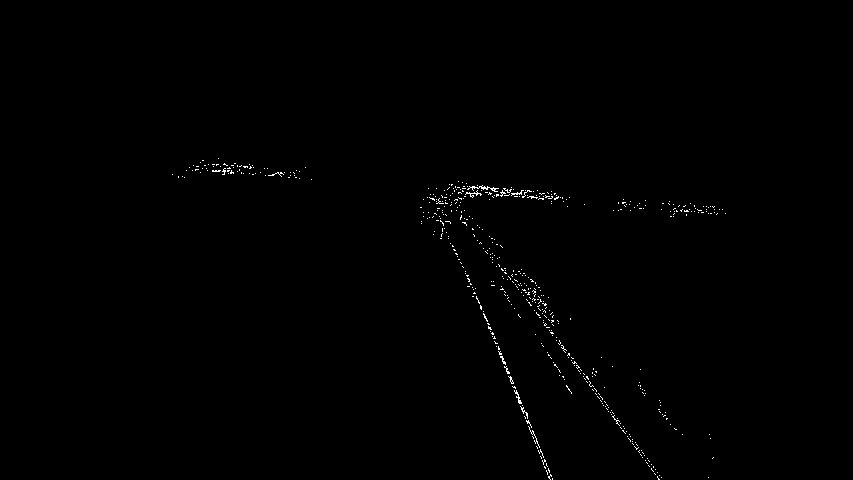}&
\includegraphics[width=0.1906\linewidth]{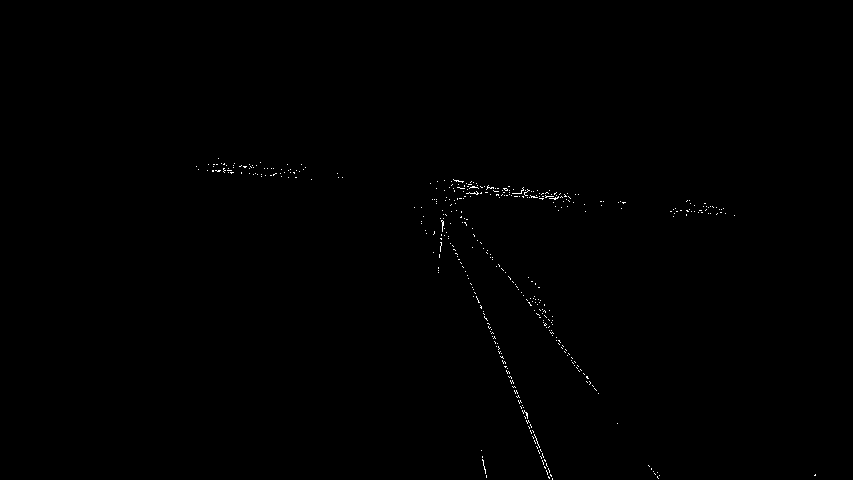}\\
Image & GT & DCNN & UNet & UCorr (ours)
\end{tabular}
\end{center}
\caption{Qualitative results for wire segmentation on our simulated flights. Each row showcases the output of various methods when applied to the input image as well as the ground truth (GT) segmentation. The visual representations are best observed in a digital format and can be examined more closely by zooming in. Our method tends to produce thinner segmentation masks, closely resembling the ground truth.}
\label{fig:qualitative_segmentation_results}
\end{figure*}

\begin{figure*}[t!]
\begin{center}
\begin{tabular}{@{}c@{\hspace{1.5mm}}c@{\hspace{1.5mm}}c@{\hspace{1.5mm}}c@{\hspace{1.5mm}}c@{}}
\includegraphics[width=0.1906\linewidth]{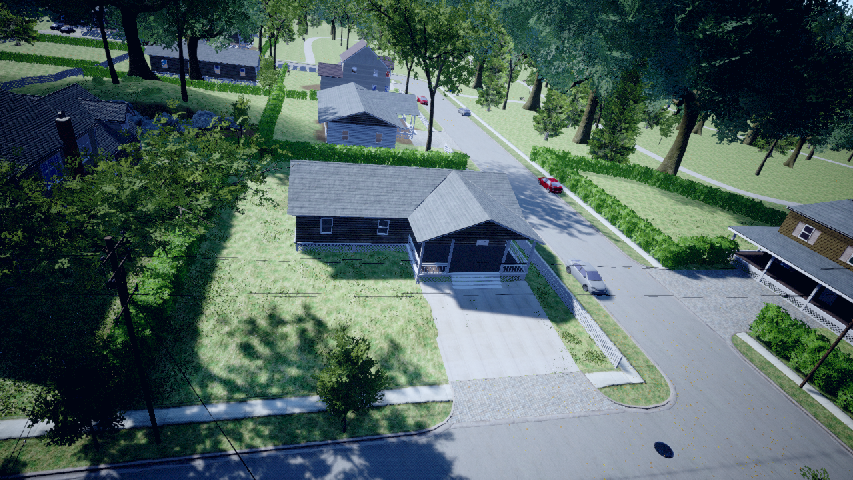}&
\includegraphics[width=0.1906\linewidth]{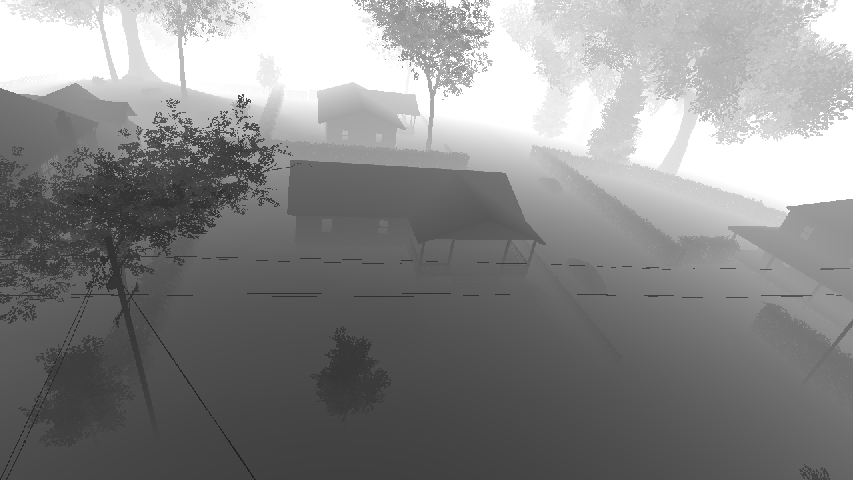}&
\includegraphics[width=0.1906\linewidth]{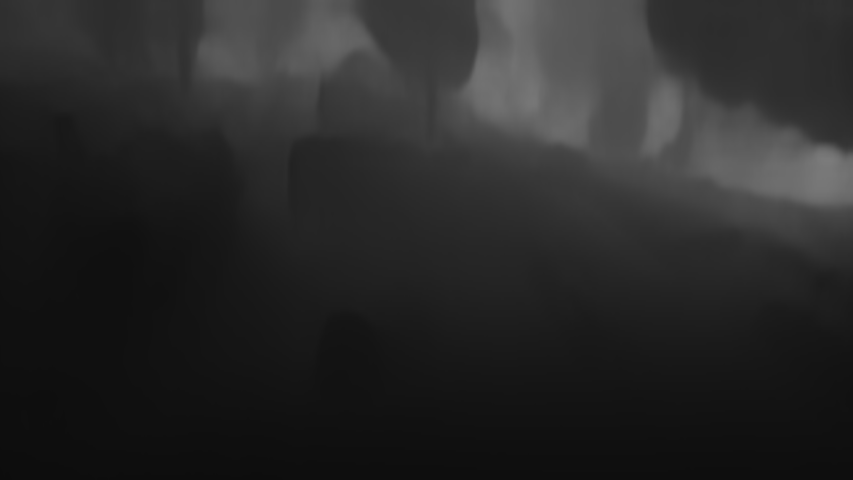}&
\includegraphics[width=0.1906\linewidth]{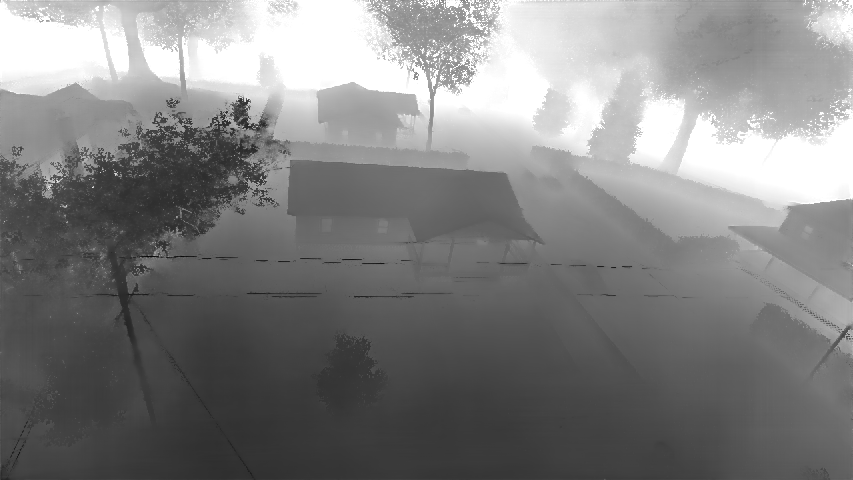}&
\includegraphics[width=0.1906\linewidth]{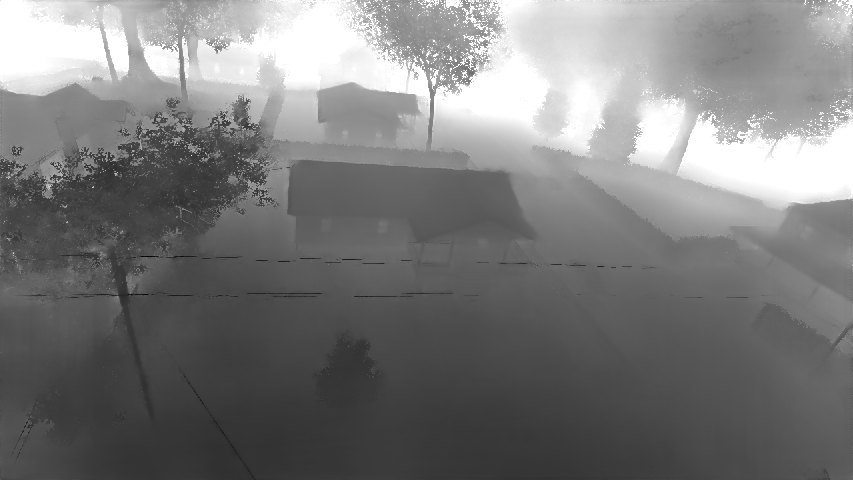}\\
\includegraphics[width=0.1906\linewidth]{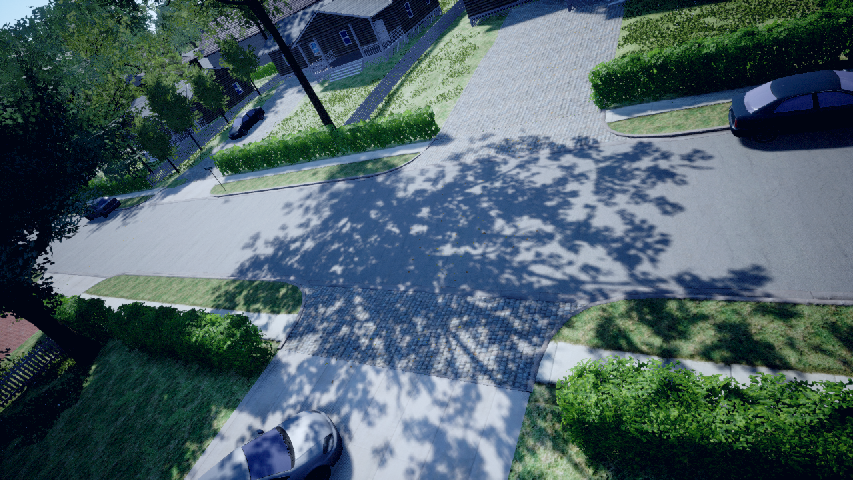}&
\includegraphics[width=0.1906\linewidth]{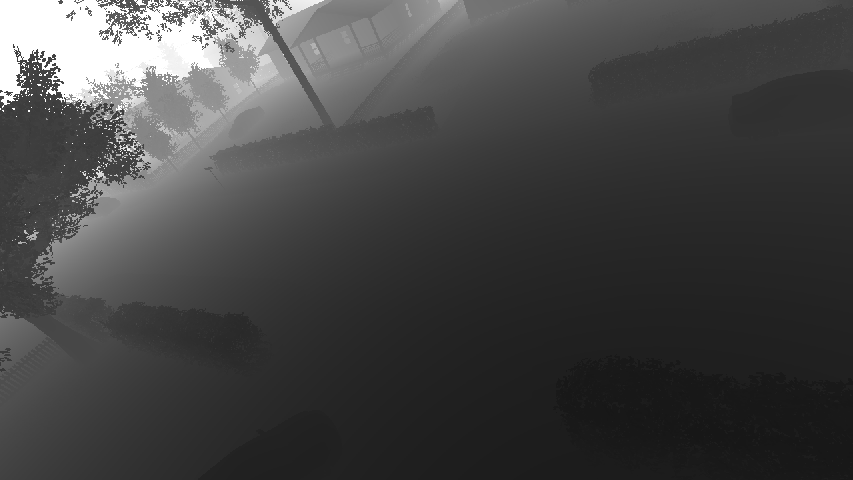}&
\includegraphics[width=0.1906\linewidth]{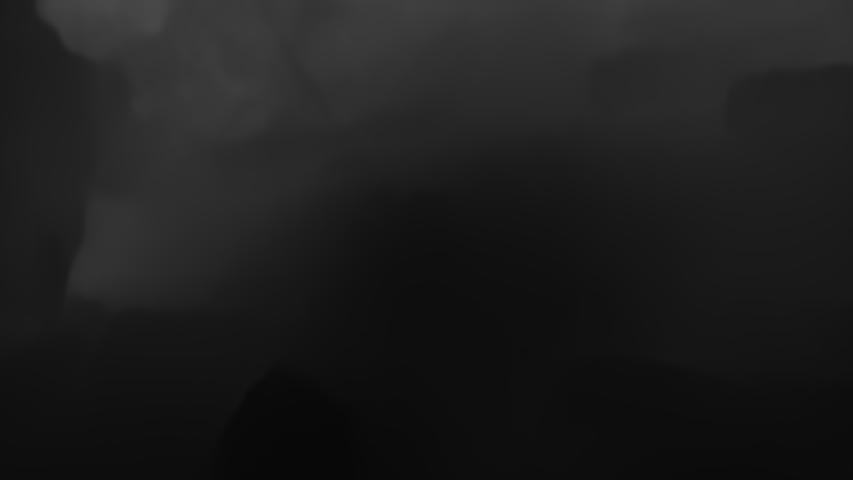}&
\includegraphics[width=0.1906\linewidth]{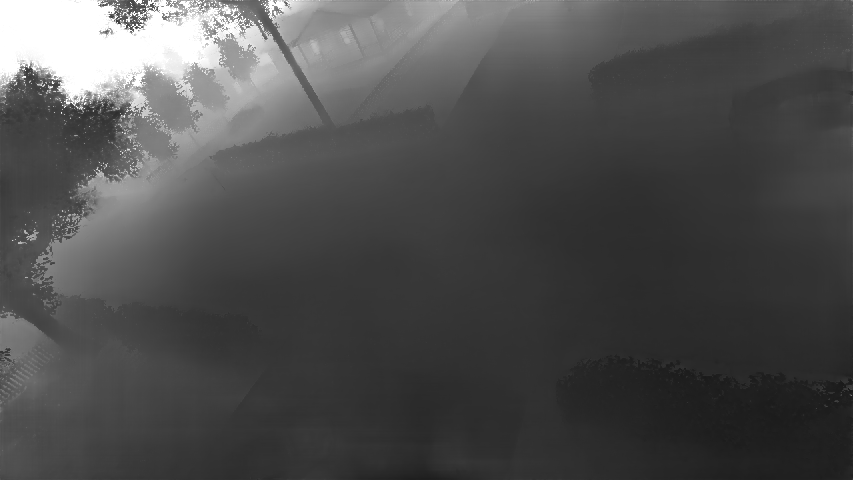}&
\includegraphics[width=0.1906\linewidth]{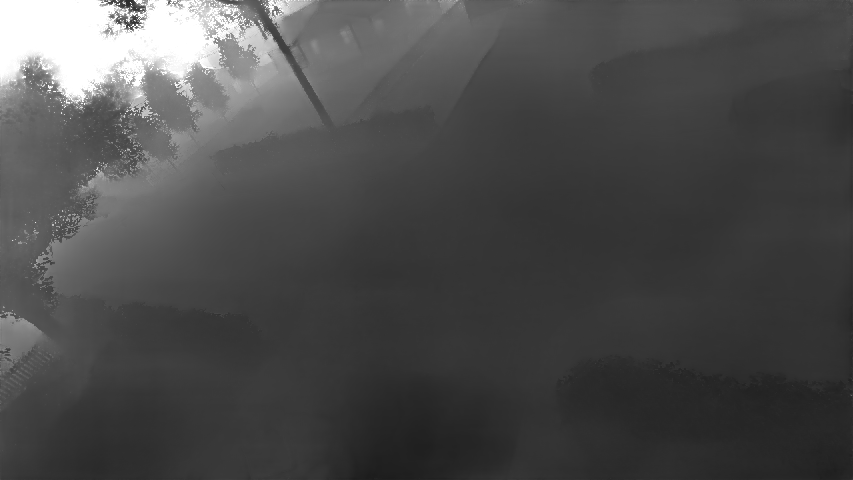}\\
\includegraphics[width=0.1906\linewidth]{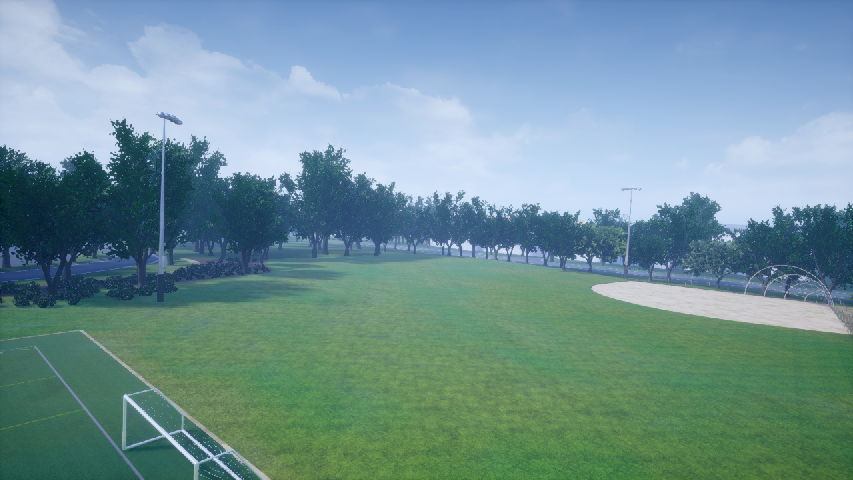}&
\includegraphics[width=0.1906\linewidth]{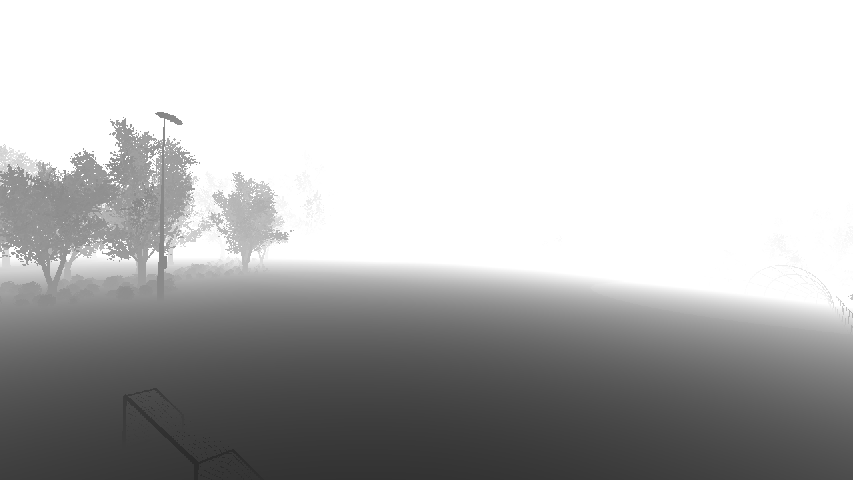}&
\includegraphics[width=0.1906\linewidth]{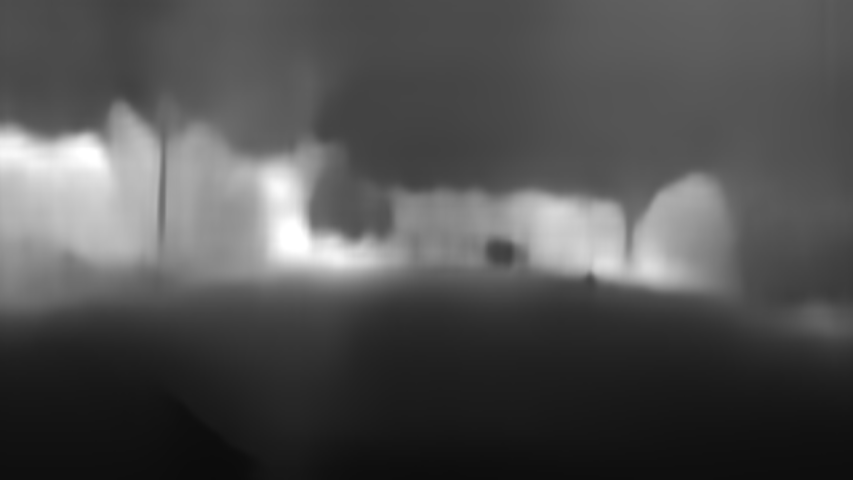}&
\includegraphics[width=0.1906\linewidth]{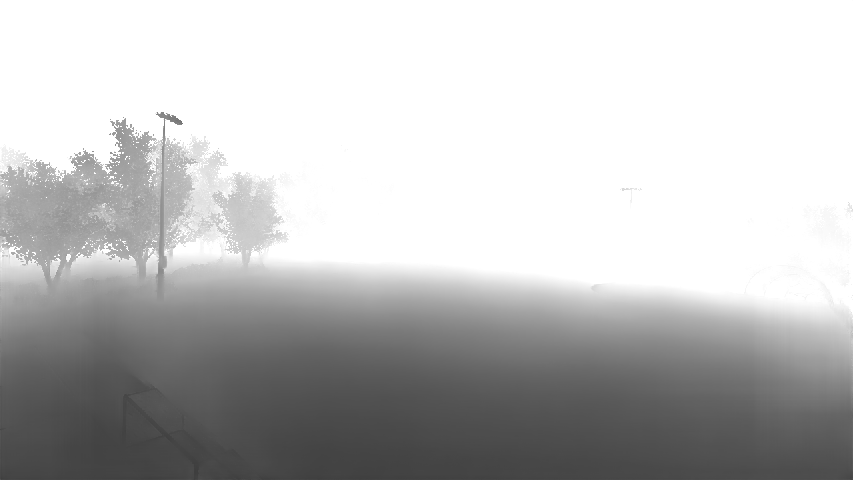}&
\includegraphics[width=0.1906\linewidth]{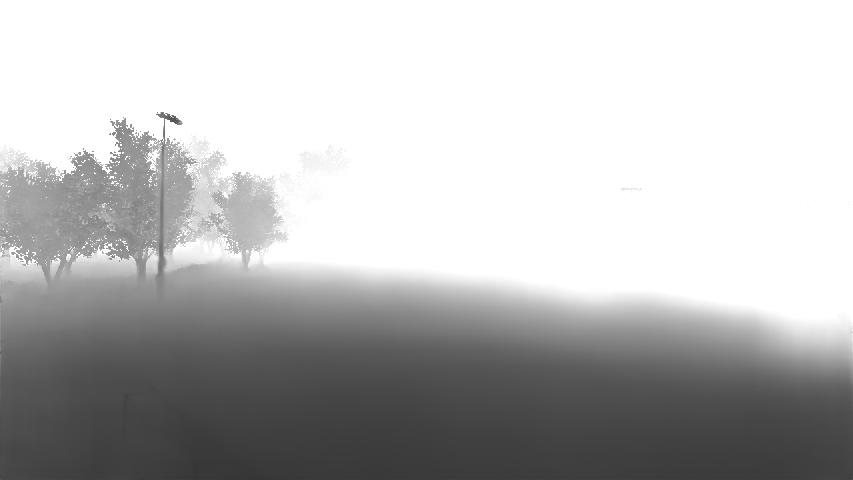}\\
Image & GT & Monodepth2 & UNet & UCorr (ours)
\end{tabular}
\end{center}
\caption{Qualitative results for depth estimation on our simulated flights. Each row showcases the output of various methods when applied to the input image, as well as the ground truth (GT). Monodepth2 is trained on KITTI and fails to generalize to the less restrictive pose and more challenging data from the simulated drone flights. In the top row, the wires are clearly visible in the depth maps for both UNet and our method.}
\label{fig:qualitative_depth_results}
\end{figure*}

\section{EXPERIMENTS}
In this section, we provide an overview of the data and metrics used for our evaluation in the joint task of wire segmentation and depth estimation. Subsequently, we present both quantitative and qualitative results of our approach in this task. Additionally, we include an ablation study to analyze the individual components of our method and their contributions to the overall performance.

\subsection{Data}
No real-world annotated data exists for both wire segmentation and precise depth estimation from aerial views. Instead, we leverage the Drone Depth and Obstacle Segmentation (DDOS) dataset~\citep{kolbeinsson2023ddos}. In total, DDOS contains 380 unique drone flights capturing \num{38 000} frames. 300 flights, are used for training while the remaining 80 flights are split evenly between validation and testing.

\subsection{Metrics} \label{section:results_metrics}
\paragraph{Segmentation metrics}
When wire detection is treated as a segmentation problem, common metrics like IoU (Intersection over Union), precision, recall, and F1 score can be used. The advantage of this approach lies in its clear objective: classifying individual pixels as wires or not.
Alternatively, some approaches involve producing best-fit lines to represent wires. These methods assess accuracy based on the distance and angle differences between proposed lines and ground truth. However, this approach can be challenging due to the non-straight nature of many wires. While various solutions exist, subtle metric variations can hinder comparisons among researchers.

Given the class imbalance (there are far fewer pixels of wires compared to not of wires), we also report the AUC (Area Under the Curve) score and average precision (AP) to evaluate the performance.

\paragraph{Depth metrics}
For depth estimation, we report the Absolute Relative Error (Abs. Rel.) and Mean Absolute Error (MAE). In addition, we introduce the Absolute Relative Error for Wire Depth (Abs. Rel. WD), a challenging metric designed for assessing models in drone-related tasks, where wires can be thin and appear free-floating.

\subsection{Training}
We train UCorr for 15 epochs on the training split of our synthetic dataset. We use a stochastic gradient descent (SGD) optimizer with momentum of 0.9 and weight decay equal to 0.01. An initial learning rate of \(5\times10^{-3}\), decaying each epoch by a factor of \(0.9\). The maximum correlation disparity is set to 10. To increase generalization, we apply augmentation to the training data comprised of: motion blur, random flips, RGB shift, color jitter, randomize hue and saturation, invert, contrast limited adaptive histogram equalization, randomize brightness and contrast and randomize gamma. Images are rescaled to \(853 \times 480\) using nearest neighbor interpolation (NNI).

\subsection{Results}

\paragraph{Quantitative results}
The quantitative results for wire segmentation and depth estimation can be found in \Cref{tab:model_comparison_synthetic_wire,tab:model_comparison_synthetic_depth}, respectively. In the wire segmentation task, our method demonstrates superior performance across all metrics, highlighting its effectiveness in accurately identifying wires. For depth estimation, our method particularly excels in the challenging absolute relative wire depth metric. This metric, which accounts for the thin nature of wires and their free-floating appearance, underscores our method's ability to accurately estimate the depth of wires in complex scenarios. 

Overall, our model excels in this joint task, demonstrating superior performance across all but one evaluated metric. This highlights the effectiveness of our approach in simultaneously addressing wire segmentation and depth estimation.

\paragraph{Qualitative results}
Qualitative wire segmentation results, including comparisons with other methods, are displayed in \Cref{fig:qualitative_segmentation_results}. In these visual examples, we can observe the effectiveness of our method in accurately segmenting wires within the images, while also assessing how it performs in comparison to other approaches. Notably, our method tends to produce thinner segmentation masks that closely resemble the ground truth, as evident in the images. Qualitative depth estimation results from our simulated flights are presented in \Cref{fig:qualitative_depth_results}. Notably, Monodepth2 struggles to generalize to drone views, given its training on KITTI \cite{Geiger2013IJRR}. Meanwhile, the visible differences between our method and UNet are minimal.

\subsection{Ablation Studies} \label{subsec:ablation}
\paragraph{Architecture Variants}
First, the influence of architectural variants on UCorr's performance is explored. As detailed in \Cref{tab:ablation_architecture}, different locations for the correlation layer within the architecture are investigated. Specifically, UCorr variants based on pixel correlation, shallow features, and deep features are compared. The results are presented in terms of relative performance compared to UCorr with deep features.
The superior performance of UCorr with deep features underscores the significance of this architectural choice.

\paragraph{Input Frames}
Next, we examine the impact of the number of input frames on the model's performance. As shown in \Cref{tab:ablation_unet_frames}, UNet's performance is assessed with varying numbers of input frames, including 1, 2, and 3 frames. The relative performance is reported concerning UNet with a single input frame, which serves as the baseline for comparison.
These findings emphasize that simply concatenating input frames is not sufficient. The correlation layer's role in integrating information across frames is a key factor in the model's success.

\paragraph{Skip-Connections}
Finally, the role of skip-connections within the UCorr architecture is evaluated. Skip-connections are known for their ability to mitigate the information bottleneck and enhance feature propagation. In  \Cref{tab:ablation_experimental}, the relative performance of UCorr without skip-connections is reported.
The results highlight the critical role of skip-connections in enhancing the model's performance. The absence of skip connections leads to performance degradation, which can be attributed to the reduced capacity for information exchange between network layers.

\begin{table}[tb!] \centering
\caption{Comparing UCorr architectural variants based on correlation layer location. UCorr (Pixel correlation) directly correlates input frame pixels. UCorr (Shallow features) employs small encoders for each input frame and correlates their shallow features. UCorr (Deep features) uses larger encoders and achieves the best performance, referred to as UCorr. Relative performance is reported compared to UCorr (Deep features).}
\begin{small}
\begin{tabular}{@{}l@{\hspace{0.4em}}cc@{\hspace{0.9em}}c@{}}\toprule
Model & \(\Delta\) Precision & \(\Delta\) Recall & \(\Delta\) F1\\ \midrule
UCorr (Pixel correlation) & -15.7\% & 3.4\% & -9.2\% \\
UCorr (Shallow features) & -31.0\% & 4.5\% & -20.6\% \\
UCorr (Deep features) & - & - & - \\
\bottomrule
\end{tabular}
\end{small}
\label{tab:ablation_architecture}
\end{table}

\begin{table}[tb!] \centering
\caption{Comparing UNet performance with different numbers of input frames. UNet (1 frame) is the default version, while (2 frames) and (3 frames) involve simple frame concatenation. Relative performance is reported with respect to UNet (1 frame), which achieves the best overall performance.}
\begin{small}
\begin{tabular}{@{}lccc@{}}\toprule
Model & \(\Delta\) Precision & \(\Delta\) Recall & \(\Delta\) F1\\ \midrule
UNet (1 frame) & - & - & - \\
UNet (2 frames) & -6.2\% & -5.1\% & -6.0\% \\
UNet (3 frames) & -1.6\% & -9.1\% & -2.8\% \\
\bottomrule
\end{tabular}
\end{small}
\label{tab:ablation_unet_frames}
\end{table}

\begin{table}[tb!] \centering
\caption{Evaluating the influence of skip-connections in UCorr on performance. Skip-connections play a crucial role in the UCorr architecture, affecting its overall performance. We report relative performance in comparison to UCorr.}
\begin{small}
\begin{tabular}{@{}lccc@{}}\toprule
Model & \(\Delta\) Precision & \(\Delta\) Recall & \(\Delta\) F1\\ \midrule
UCorr & - & - & - \\
w/o skip-connections & -92.6\% & 15.6\% & -88.5\% \\ 
\bottomrule
\end{tabular}
\end{small}
\label{tab:ablation_experimental}
\end{table}

\section{DISCUSSION}

One limitation of our approach is its dependency on exactly two sequential input frames for temporal fusion. While this method is effective in many scenarios, it presents a challenge when the drone is stationary or moving slowly, as there may be minimal discernible differences between consecutive frames. It would be advantageous if our method could adapt to varying frame numbers or capture and store scene flow during drone movement, addressing these situations more effectively.

A significant limitation is the absence of real-world data tailored for wire detection and depth estimation. Real-data testing is currently unfeasible as no such datasets exist, limiting the assessment of our method's real-world applicability.
While testing against a broader range of benchmarks, especially those outside of methods tailored for wire detection, would have been beneficial, our work is constrained by computational resources.

This research also raises potential security and dual-use concerns, as the technology could be applied both for legitimate purposes and, in some cases, malicious applications. Researchers must remain vigilant in addressing these concerns and promoting the responsible and secure use of their findings.

The natural next step is implementing our method on a real drone, in collaboration with a dedicated hardware team. This practical deployment will validate the efficacy of our approach in real-world scenarios and pave the way for further enhancements based on empirical results. 
Additionally, exploring novel knowledge distillation techniques \citep{miles2023closer} offers opportunities to develop smaller, more efficient models.

\section{CONCLUSION}

Our contributions represent three significant advancements in the field. Firstly, we illuminate the underexplored domain of wire detection and depth estimation, recognizing its growing importance in applications like autonomous navigation and infrastructure maintenance. Secondly, our introduction of UCorr, an innovative model tailored for monocular wire segmentation and depth estimation, not only outperforms existing methods but also sets a valuable benchmark for the field. Finally, our novel wire depth evaluation metric enhances evaluation precision and comprehensiveness. Collectively, our work serves as a pivotal point for future research in this domain, opening doors for innovation and improved solutions.

\newpage
\bibliographystyle{apalike}
{\small
\bibliography{refs}}

\end{document}